\documentclass[letterpaper, 10 pt, conference]{ieeeconf}  

\IEEEoverridecommandlockouts                              
                                                          
\usepackage{graphicx} 

\usepackage{algorithm}
\usepackage{algorithmic}

\usepackage{amssymb}
\usepackage{amsmath}    
\usepackage{multirow}

\usepackage{booktabs}  
\usepackage{graphicx}
\usepackage{subcaption}  
\title{\LARGE \bf
SURE: Semi-dense Uncertainty-REfined Feature Matching
}

\author{Sicheng Li$^{1,3}$, Zaiwang Gu$^{2}$, Jie Zhang$^{3}$, Qing Guo$^{4}$, Xudong Jiang$^{1}$ and Jun Cheng$^{2*}$
\thanks{*Corresponding author: Jun Cheng ({cheng\_jun@a-star.edu.sg}).}
\thanks{$^{1}$S. Li and X. Jiang are with the School of Electrical and Electronic Engineering, 
Nanyang Technological University (NTU), Singapore.}
\thanks{$^{2}$J. Cheng and Z. Gu are with the Institute for Infocomm Research (I$^2$R), 
Agency for Science, Technology and Research (A*STAR), Singapore.}
\thanks{$^{3}$J. Zhang and S. Li are with Institute of High Performance Computing, Agency for Science, Technology and Research (A*STAR), Singapore.}
\thanks{$^{4}$Q. Guo is  with the College of Computer Science, 
Nankai University, Tianjin, China.}
\thanks{This research is supported by the National Research Foundation Singapore under the AI Singapore Programme (NO. AISG4-GC-2023-008-1B) and A*STAR under its MTC Programmatic Funds (GranNo. M23L7b0021).}%
}
\begin{document}

\maketitle
\thispagestyle{empty}
\pagestyle{empty}

\begin{abstract}
Establishing reliable image correspondences is essential for many robotic vision problems. However, existing methods often struggle in challenging scenarios with large viewpoint changes or textureless regions, where incorrect correspondences may still receive high similarity scores. This is mainly because conventional models rely solely on feature similarity, lacking an explicit mechanism to estimate the reliability of predicted matches, leading to overconfident errors. To address this issue, we propose SURE, a Semi-dense Uncertainty-REfined matching framework that jointly predicts correspondences and their confidence by modeling both aleatoric and epistemic uncertainties. Our approach introduces a novel evidential head for trustworthy coordinate regression, along with a lightweight spatial fusion module that enhances local feature precision with minimal overhead. We evaluated our method on multiple standard benchmarks, where it consistently outperforms existing state-of-the-art semi-dense matching models in both accuracy and efficiency. our code will be available on https://github.com/LSC-ALAN/SURE.

\end{abstract}


\section{INTRODUCTION}

Feature matching aims to establish visual correspondences between two images and serves as a cornerstone in many robotic vision applications, including Structure-from-Motion (SfM)~\cite{lindenberger2021pixel}, visual localization~\cite{sarlin2019coarse}, 3D reconstruction~\cite{heinly2015reconstructing}, and SLAM~\cite{mur2017orb,Orbeez-slam}.  However, achieving accurate and reliable feature matching remains challenging in real-world scenarios due to factors such as repetitive textures,   occlusions, and large viewpoint, scale or modality changes ~\cite{yepeng2025tip},~\cite{liu2025modality},~\cite{jingbo2025}.

Recent methods such as LoFTR~\cite{sun2021loftr} abandon explicit keypoint detection and adopt a coarse-to-fine framework. Using transformer~\cite{vaswani2017attention} as feature extractor, they compute patch-wise similarities for coarse matching, followed by fine-level refinement at the pixel level. Building on this idea, MatchFormer~\cite{wang2022matchformer} incorporates multiscale features to improve robustness across varying resolutions, while E-LoFTR~\cite{wang2024efficient} introduces more efficient modules to optimize fine-level matching.

\begin{figure}[t]
  \centering
  \begin{subfigure}[t]{0.45\textwidth}
    \centering
    \includegraphics[width=\linewidth]{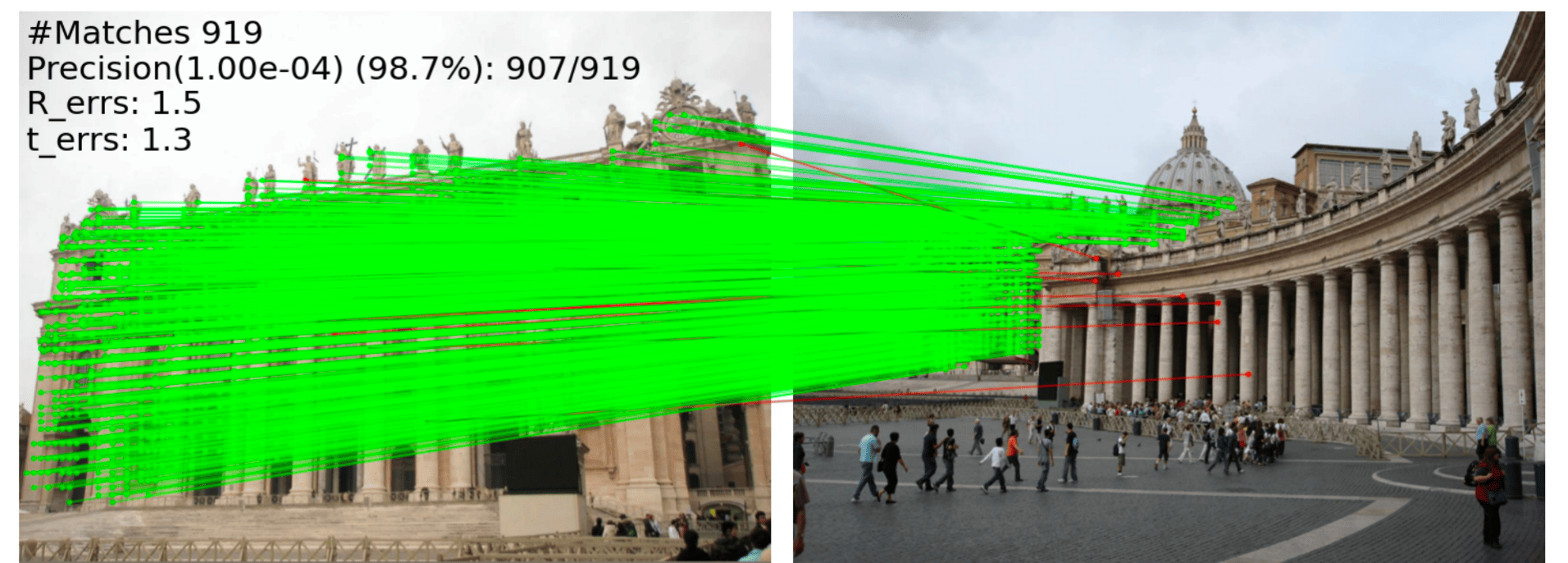}
    \caption{SURE}
  \end{subfigure}
  \hfill
  \begin{subfigure}[t]{0.45\textwidth}
    \centering
    \includegraphics[width=\linewidth]{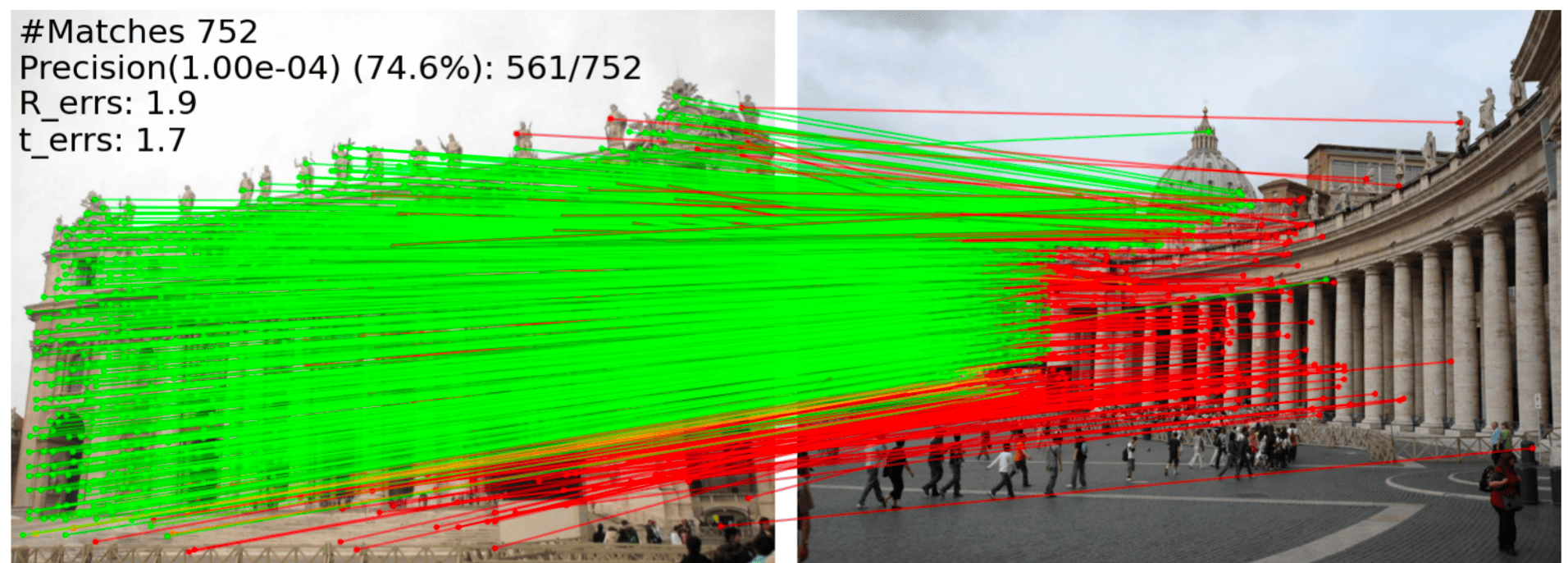}
    \caption{E-LoFTR~\cite{wang2024efficient}}
  \end{subfigure}
  
  \caption{Comparison of E-LoFTR and our method on the MegaDepth dataset. Lines highlighted in green and red correspond to points with an epipolar error less than or greater than $10^{-4}$ respectively. }
  \label{fig1}
\end{figure}

Despite their success, most of the existing methods still suffer from two critical limitations. First, most existing methods estimate confidence purely based on feature similarity, without explicitly modeling the intrinsic trustworthiness of the prediction itself. This becomes problematic in challenging scenarios such as large viewpoint changes or textureless regions, where incorrect matches often receive high similarity scores, and thus cannot be effectively filtered. In Structure-from-Motion or SLAM systems, unreliable matches can severely affect pose estimation and 3D reconstruction~\cite{engelmann2021points}, making the ability to assess match reliability just as important as finding the matches. Second, many existing models prioritize accuracy at the cost of efficiency, relying on large architectures and highly complex computations, which limit their applicability in real-time or resource-constrained scenarios.

To address these issues, we propose SURE, a Semi-dense Uncertainty-Refined matching framework that jointly estimates correspondences and their associated uncertainties. At the core of SURE is trustworthy regression, which employs evidential learning\cite{amini2020deep} to predict both aleatoric and epistemic uncertainties along with the correspondence coordinates. Instead of computing similarity across high-resolution 2D windows, SURE estimates two sets of 1D heatmaps representing the marginal distributions of the coordinates $x$ and $y$. These 1D heatmaps are treated as probabilistic evidence, enabling the model to jointly infer precise coordinates and associated uncertainties in a principled and efficient manner. This formulation not only reduces computational complexity significantly, but also yields a principled confidence estimate for each match. To further enhance fine-level prediction, we introduce a lightweight spatial fusion strategy that effectively preserves detailed low-level spatial information, thereby improving accuracy without incurring significant computational overhead.
As illustrated in Figure~\ref{fig1}, our method yields a higher proportion of geometrically consistent correspondences, with noticeably fewer mismatches.

We evaluated our method on several widely used benchmarks, including MegaDepth~\cite{li2018megadepth} and ScanNet~\cite{dai2017scannet}, where SURE consistently outperforms existing state-of-the-art semi-dense matching models in both accuracy and efficiency. Our approach demonstrates not only strong generalization across scenes, but also the ability to produce trustworthy confidence scores for match filtering and downstream tasks.

Our main contributions are summarized as follows.
\begin{itemize}
\item We propose SURE, a novel semi-dense matching framework that integrates correspondence prediction with uncertainty estimation.

\item We introduce an evidential regression head that jointly models aleatoric and epistemic uncertainties, providing reliable confidence scores for match evaluation.

\item We propose a spatial fusion module that refines local features by incorporating hierarchical spatial information and enhancing structural details.

\item We demonstrate that our method surpasses previous state-of-the-art approaches such as E-LoFTR in both accuracy and efficiency on standard benchmarks.

\end{itemize}

\section{RELATED WORKS}
\subsection{Feature Matching}
\subsubsection{Sparse Matching}
Sparse feature matching methods identify and describe a limited set of keypoints. Early approaches such as SIFT~\cite{lowe2004distinctive} and ORB~\cite{rublee2011orb} rely on handcrafted descriptors with nearest-neighbor matching. With deep learning, methods like R2D2~\cite{r2d2} and SuperPoint~\cite{detone2018superpoint} learn robust detectors and descriptors using convolutional networks. Extensions include semantic-aware SLAM frameworks that enhance correspondence and loop closure~\cite{10610238}. SuperGlue~\cite{sarlin2020superglue} models relationships between features via graph neural networks, while LightGlue~\cite{lindenberger2023lightglue} improves efficiency. Descriptor distillation~\cite{guo2023descriptor} learns compact representations through teacher–student training, and LiftFeat~\cite{liu2025liftfeat} integrates pseudo-3D geometric cues to improve robustness. Despite these advances, sparse methods remain limited in low-texture or repetitive regions due to unreliable keypoint detection.

\subsubsection{Dense Matching}
Dense matchers aim to predict correspondences across nearly all image pixels. Early approaches such as DGC-Net~\cite{melekhov2019dgc} and DRC-Net~\cite{li20dualrc} rely on 4D correlation volumes to explore full matching spaces, while PDC-Net~\cite{pdcnet} improves performance through progressive refinement with deformable convolutions. More recent methods, including DKM~\cite{edstedt2023dkm}, formulate dense matching probabilistically and predict confidence maps to filter unreliable matches. RoMa~\cite{edstedt2024roma} further leverages frozen DINOv2~\cite{oquab2024dinov} features with a dedicated convolutional decoder for refinement. Despite strong accuracy and alignment, dense methods remain computationally expensive due to high-resolution processing and large feature volumes.


\subsubsection{Semi-Dense Matching}
Semi-dense methods aim to strike a balance between matching density and computational cost. LoFTR~\cite{sun2021loftr} pioneers a coarse-to-fine strategy, first building coarse-level correspondences on a downsampled grid, then refines them using fine-grained feature patches. This approach enables more complete coverage than sparse techniques while avoiding the cost of full-resolution matching. MatchFormer~\cite{wang2022matchformer} leverages a hierarchical transformer to jointly model global context. 
TopicFM~\cite{giang2023topicfm} introduces topic-based modules to leverage semantic context for more robust matching, and 
E-LoFTR~\cite{wang2024efficient} addresses efficiency limitation by compressing local features and refining matches in a more compact representation.
However, challenges remain in ensuring reliable fine-level alignment without high overhead. 
Our method follows this semi-dense paradigm to maintain wide coverage and efficiency, but differs by integrating uncertainty-aware predictions for more reliable refinement.

\begin{figure*}[ht]
\centering
\includegraphics[width=1\textwidth]{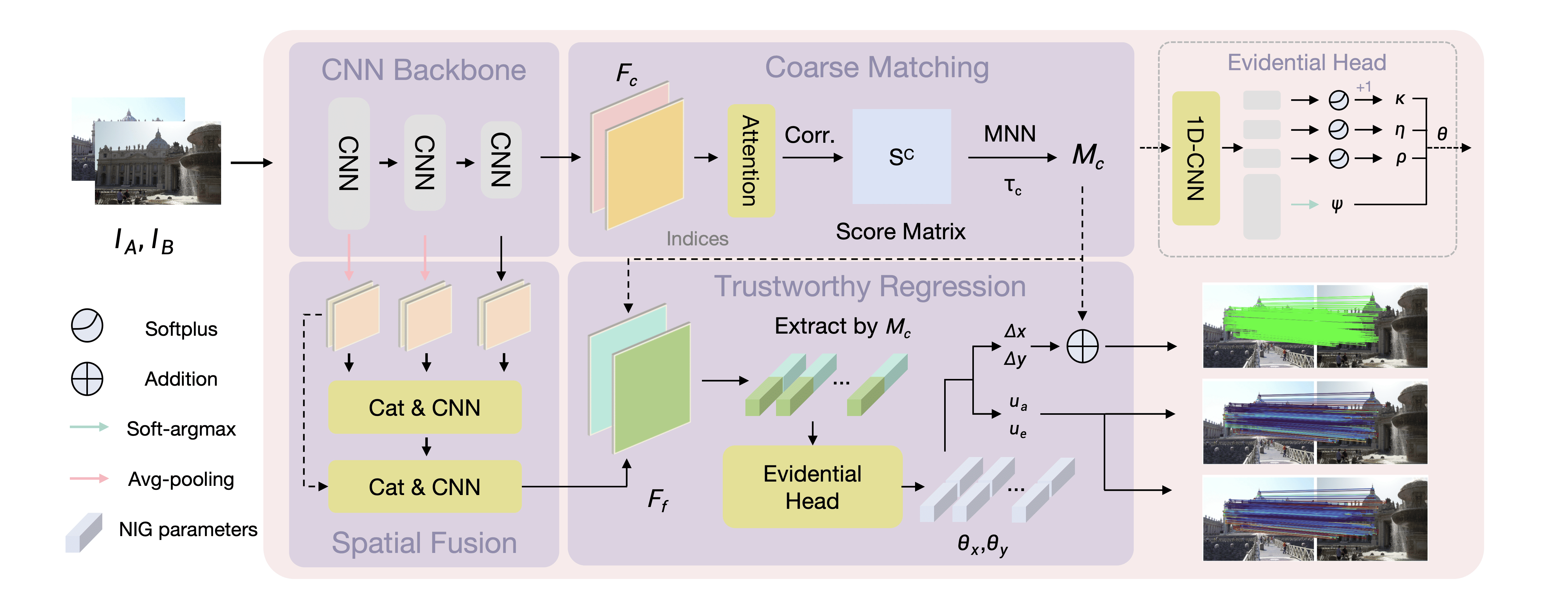} 
\caption{Overview of the proposed SURE framework. (1) A backbone extracts coarse features $F_c$ and a Spatial Fusion Module provides fine features $F_f$. (2) The coarse matching module produces initial correspondences $M_c$, which are used to sample $F_f$ for fine-level refinement. (3) Trustworthy Regression produces precise offsets $(\Delta x, \Delta y)$ along with uncertainty estimates. Specifically, an Evidential Head predicts the parameters $(\psi, \eta, \kappa, \rho)$ of a Normal-Inverse-Gamma distribution, which jointly encode the offset and its associated aleatoric and epistemic uncertainties.}
\label{fig2}
\end{figure*}

\subsection{Uncertainty Estimation}
Recent advances in deep learning have led to various uncertainty modeling techniques. Bayesian neural networks~\cite{NIPS2017_2650d608} and deep ensembles~\cite{Wen2020BatchEnsemble} model epistemic uncertainty by treating weights as distributions or training multiple models, but are often computationally expensive. Monte Carlo Dropout~\cite{bayesian2016} offers a simpler approximation via stochastic forward passes. Deep Evidential Learning~\cite{sensoy2018evidentialdeeplearningquantify,NEURIPS2018_3ea2db50} avoids sampling by using deterministic evidence-based inference. For regression, Deep Evidential Regression~\cite{amini2020deep} extends this framework by modeling outputs with a normal-inverse gamma prior, allowing joint estimation of aleatoric and epistemic uncertainties in a single pass.

In correspondence estimation, uncertainty modeling remains to be explored. Evidential learning has recently been applied to other tasks such as stereo matching~\cite{Lou_2023_ICCV} and vision language calibration~\cite{oh2024towards}, but its potential in feature matching has not yet been fully explored.
To address this gap, we introduce an evidential formulation that jointly estimates correspondence offsets and quantifies both aleatoric and epistemic uncertainties in a principled manner.

\section{METHOD}
This section presents an overview of the proposed SURE framework. As depicted in Figure ~\ref{fig2}, the architecture consists of four main components: a CNN backbone for feature extraction, a coarse matching module to generate initial correspondence candidates via global context-aware features, a lightweight spatial fusion module that integrates multi-scale details to enrich local features, and a trustworthy regression stage incorporating an evidential head that predicts precise matches along with aleatoric and epistemic uncertainty estimates. 
This design enables SURE to provide accurate, reliable, and efficient semi-dense matching by combining hierarchical feature representations and principled uncertainty modeling.

\subsection{Feature Extraction}
To extract hierarchical visual representations, we utilize a single-branch compact backbone with RepVGG~\cite{ding2021repvgg}. Given a pair of input images $I^A, I^B \in \mathbb{R}^{C \times H \times W}$, the network generates multilevel features, where $C$, $H$, and $W$ refer to the number of channels, height, and width, respectively. The deepest layer produces coarse descriptors $F_c^A, F_c^B \in \mathbb{R}^{C_c \times H_c \times W_c}$, where $H_c = \frac{H}{8}$, $W_c = \frac{W}{8}$, and $C_c = 256$, taking advantage of broad contextual patterns.

\subsection{Coarse Matching}
Following prior work~\cite{wang2024efficient}, we apply self and cross attention to the coarse-level features $F_c^A$and$ F_c^B$,
yielding the enhanced representations $ \hat{F}^A_c $ and $ \hat{F}^B_c $. We establish initial correspondences by computing a bidirectional similarity matrix followed by confidence-based filtering.

We first calculate the coarse similarity matrix \( S^c \in \mathbb{R}^{H_cW_c \times H_cW_c} \) using the inner product scaled by temperature:
\begin{equation}
S^c_{i,j} = \frac{1}{\tau} \cdot \langle \hat{F}^A_c(i), \hat{F}^B_c(j) \rangle,
\end{equation}
where \( \tau \) is a fixed scalar that controls the sharpness of the distribution.

To infer the matching confidence in both directions, we perform softmax normalization along rows and columns, respectively:
\begin{equation}
\begin{aligned}
P^{A \rightarrow B}_{i,j} = \frac{\exp(S^c_{i,j})}{\sum_{k} \exp(S^c_{i,k})}, 
P^{B \rightarrow A}_{i,j} = \frac{\exp(S^c_{i,j})}{\sum_{k} \exp(S^c_{k,j})}.
\end{aligned}
\end{equation}

Following common practice, we adopt mutual nearest neighbor (MNN) filtering~\cite{lindenberger2023lightglue}, selecting correspondences that are bidirectional maxima and exceed a confidence threshold~$\tau_c$. The resulting set of coarse matches, denoted as~$\mathcal{M}_c$, serves as candidate correspondences for fine-level refinement.

\subsection{Spatial Fusion Module}
Unlike traditional FPN-style designs in feature matching~\cite{sun2021loftr}, which upsample low-resolution features through top-down fusion, we adopt a streamlined multi-scale fusion strategy optimized for fine stage regression. While prior methods often restore features to full resolution for dense refinement, leading to high computational cost, our method uniformly aligns all features to a fixed resolution $\frac{1}{8}$ for the final regression. This avoids costly cropping and maintains efficiency. To retain structural detail, we incorporate a high-resolution enhancement path inspired by HRNet~\cite{wang2020deep}, enriching the fused features with both spatial precision and semantic depth.

Specifically, we extract intermediate features from the backbone at three spatial scales: $\left\{ \frac{1}{2}, \frac{1}{4}, \frac{1}{8} \right\}$ of the input resolution. These features are denoted as $\{ F_{(s)} \}_{s \in \{2,4,8\}}$, where $F_{(s)} \in \mathbb{R}^{C^{(s)} \times \frac{H}{s} \times \frac{W}{s}}$ corresponds to the feature map at scale $\frac{1}{s}$. To unify the dimensions of the channels, each $F_{(s)}$ is projected onto a common embedding space of size $C_f = 256$ using a $1 \times 1$ convolution.


The projected features are then downsampled through adaptive average pooling to match the spatial resolution of the $\frac{1}{8}$ scale, resulting in aligned features $\left\{ \hat{F}_{\frac{1}{2}}, \hat{F}_{\frac{1}{4}}, \hat{F}_{\frac{1}{8}} \right\} \in \mathbb{R}^{C_f \times \frac{H}{8} \times \frac{W}{8}}$. These are concatenated along the channel dimension and fused through a $1 \times 1$ convolution followed by batch normalization and ReLU activation, producing the fused representation $F_{\text{fused}}$.

To further preserve high-frequency details, we introduce a residual path from the original $F_{\frac{1}{2}}$ scale. It is processed through a separate $1 \times 1$ convolution and pooled to the same $\frac{1}{8}$ resolution. The final fine-level feature $F_f$ is computed as:
\begin{equation}
F_f = F_{\text{fused}} + \mathrm{Pool} \left( \mathrm{Conv}_{1\times1}(\hat{F}_{\frac{1}{2}}) \right).
\end{equation}
The resulting fused representation $F_f \in \mathbb{R}^{C_f \times H_\frac{1}{8} \times W_\frac{1}{8}}$ retains both the semantic context and local structural details, which benefits downstream fine-level refinement and uncertainty estimation.

\begin{figure*}[t]
\centering
\includegraphics[width=1\textwidth]{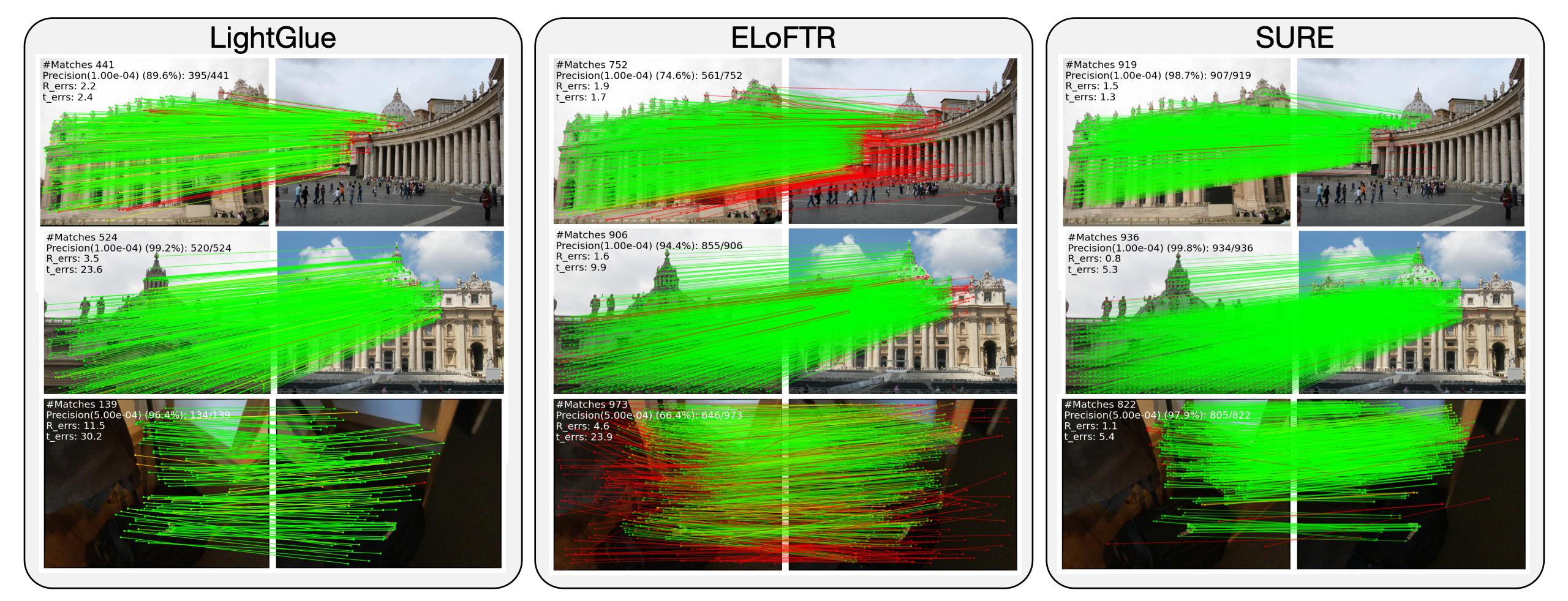} 
\caption{Qualitative comparisons of SURE against Light Glue and E-LoFTR on both indoor and outdoor scenes. SURE achieves a higher number of correct matches and reduces mismatches, demonstrating robustness in low-texture areas as well as under significant viewpoint and lighting variations. Red regions denote points with epipolar error exceeding $5 \times 10^{-4}$ for indoor and $1 \times 10^{-4}$ for outdoor scenes.}
\label{fig3}
\end{figure*}

\begin{figure}[t]
  \centering
  \begin{subfigure}[t]{0.40\textwidth}
    \centering
    \includegraphics[width=\linewidth]{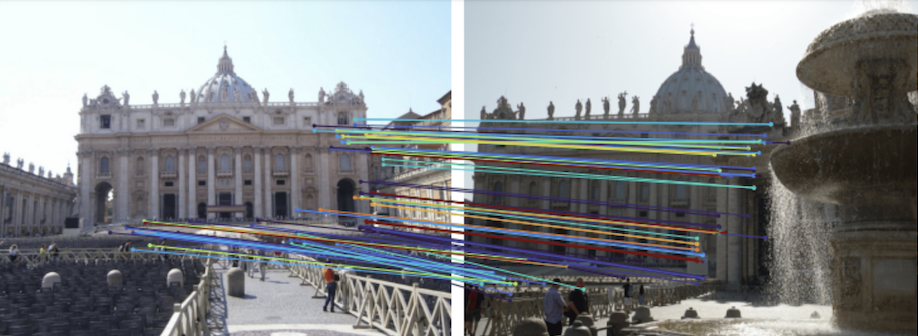}
    \caption{epistemic uncertainty}
  \end{subfigure}
  \hfill
  \begin{subfigure}[t]{0.40\textwidth}
    \centering
    \includegraphics[width=\linewidth]{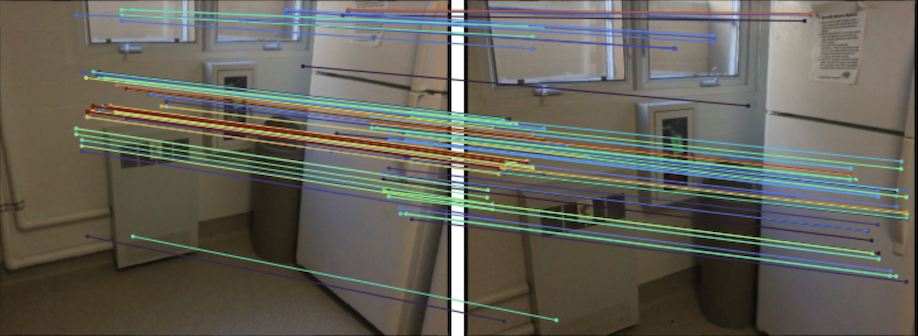}
    \caption{aleatoric uncertainty}
  \end{subfigure}

  \caption{We selected large viewpoint changes and weak-texture scenarios. Among 2048 correspondences, the 50 pairs with the highest model uncertainty and data uncertainty were chosen. The lighter the line color, the higher the uncertainty.}
  \label{fig4}
\end{figure}

\subsection{Trustworthy Regression}


\subsubsection{Probabilistic Modeling With Uncertainty}
From an evidential perspective, each offset $ z $ is sampled from a Gaussian distribution with unknown parameters $\xi$ and $\upsilon^2$.
These parameters are further modeled by placing a normal prior in $\xi$ and an inverse gamma prior in $\upsilon^2$.
\begin{equation}
z \sim \mathcal{N}(\xi, \upsilon^2),\xi \sim \mathcal{N}(\psi, \upsilon^2 \eta^{-1}),
\upsilon^2 \sim \mathcal{IG}(\kappa, \rho),
\end{equation}
where $\mathcal{IG}$ denotes the inverse gamma distribution, $\psi \in \mathbb{R}$, $\eta > 0$, $\kappa > 1$, and $\rho > 0$.
Assuming the independence between the mean and the variance, the joint posterior
$q(\xi, \upsilon^2)$ follows a normal-inverse-gamma (NIG) distribution, parameterized by $\theta=(\psi, \eta, \kappa, \rho)$:
\begin{equation}
q(\xi, \upsilon^2) = \mathcal{NIG}(\psi, \eta, \kappa, \rho).
\end{equation}

The expected prediction $\hat{z}$, along with its aleatoric uncertainty $u_a^z$ and epistemic uncertainty $u_e^z$, is derived from the offset distribution, whose mean represents the predicted offset, and variance encodes both types of uncertainty, following the formulation in~\cite{amini2020deep}.

\begin{equation}
\hat{z} = \psi , u_a^z = \frac{\rho}{\kappa - 1},
u_e^z = \frac{\rho}{\eta (\kappa - 1)}.
\end{equation}


To train this evidential model, we minimize a loss composed of two terms: a negative log-evidence term derived from the NIG distribution.
\begin{align}
\mathcal{L}^{\text{evi}}(\mathbf{w}) =\  
& \frac{1}{2}\log \left(\frac{\pi}{\eta} \right) 
- \kappa \log(\Theta) \nonumber \\
& + \left(\kappa + \frac{1}{2} \right) 
    \log\left( (\mathbf{y}^* - \psi)^2 \eta + \Theta \right) \nonumber \\
& + \log \left( \frac{\Gamma(\kappa)}{\Gamma(\kappa + \frac{1}{2})} \right)
\label{eq:evi-loss},
\end{align}
where $\Theta = 2 \rho (1 + \eta)$ and $\mathbf{w}$ denotes the network-estimated parameters.

A regularization term is introduced to penalize incorrect predictions with high confidence:
\begin{equation}
\mathcal{L}^{\text{reg}}(\mathbf{w}) = |\mathbf{y}^* - \psi| \cdot \Phi, \quad \text{with} \quad \Phi = 2\eta + \kappa,
\end{equation}
where $\mathbf{y}^*$ is the ground truth label of offsets, and $\Phi$ quantifies total evidence.

The total uncertainty-aware loss is computed over all $N$ predictions as:

\begin{equation}
\mathcal{L}_f(\mathbf{w}) = \frac{1}{N} \sum_{j=1}^{N} \left( \mathcal{L}^{\text{evi}}_j(\mathbf{w}) + \zeta \mathcal{L}^{\text{reg}}_j(\mathbf{w}) \right),
\end{equation}
where $\zeta > 0$ is the regularization coefficient.

\subsubsection{Evidential Regression Head}

Inspired by coordinate classification approaches~\cite{li2022simcc}, we design a lightweight 1D regression head that decouples offset estimation along the $x$ and $y$ axes, allowing for efficient and stable sub-pixel refinement.

Given a set of coarse matches, we first extract their corresponding fine-level descriptors from $F^A_f$ and $F^B_f$, and concatenate them to form a fused feature tensor $\hat{F}_f \in \mathbb{R}^{M \times 2d}$, where $M$ denotes the number of coarse matches and $d$ is the dimensionality of each feature vector. This tensor serves as the input to two independent regression branches for each axis.

The vector $\hat{F}_f$ is subsequently processed by two independent 1-D convolutional regression heads, one for each axis. Although both heads take the same input $\hat{F}_f$, they have separate parameters and output structures.

Each head produces a $(N + 3)$-dimensional output vector that directly parameterizes a Normal-Inverse-Gamma (NIG) distribution. Specifically, the first $N$ elements are spatial logits, from which a soft-argmax operation computes the expected value $\psi \in [-0.5, 0.5]$, which also serves as the predicted offset $z$ along the corresponding axis.
The remaining three values are transformed through soft plus activation to obtain the additional NIG parameters: $\eta$, $\kappa$, and $\rho$, which together characterize the predictive uncertainty of the model. This design allows the model to jointly learn both the prediction and its associated confidence in a fully differentiable and probabilistic manner.

\subsubsection{Uncertainty Filtering}
Given the predicted parameters of the NIG distribution, the aleatoric uncertainty $u_a$, and the epistemic uncertainty $u_e$ along both axes can be calculated using Eq.(6). For each matched pair, we obtain $(u_a^x, u_a^y)$ and $(u_e^x, u_e^y)$ from the respective regression heads. The final aleatoric and epistemic uncertainties are obtained by averaging the values across axes:
\begin{equation}
u_a = \frac{u_a^x + u_a^y}{2}, \quad u_e = \frac{u_e^x + u_e^y}{2}.
\end{equation}

We then filter the predicted matches based on these uncertainty estimates. Specifically, we set thresholds $\tau_a$ and $\tau_e$, and discard any predictions whose $u_a > \tau_a$ or $u_e > \tau_e$. These thresholds are chosen as quantiles to retain a desired percentage of the most confident predictions.

\subsection{Loss}

We supervise the matching process at both the coarse and fine levels to ensure robust global localization and subpixel precision.

\subsubsection{Coarse-Level Loss} 
Following previous work~\cite{lu2025jamma}, we adopt a dual-direction focal loss to supervise the coarse matching probability matrices $ P^{A \rightarrow B} $ and $ P^{B \rightarrow A} $, using ground truth correspondences $ P_c^{gt} $ generated by warping the grid centers via depth and camera pose.

\begin{equation}
\mathcal{L}_c = \text{FL}(P_c^{gt}, P^{A \rightarrow B}) + \text{FL}(P_c^{gt}, P^{B \rightarrow A}),
\end{equation}
where FL denotes the focal loss~\cite{lin2017focal} defined as:

\begin{equation}
\text{FL}(p) = - \alpha (1 - p)^\gamma \log p,
\end{equation}
and $ \alpha $, $ \gamma $ are the weighting and focusing parameters, respectively.

\subsubsection{Total Loss} 
The final training objective is a weighted sum of the two terms:

\begin{equation}
\mathcal{L} = \lambda_c \mathcal{L}_c + \lambda_f( \mathcal{L}_f(\mathbf{w}^x)+\mathcal{L}_f(\mathbf{w}^y)),
\end{equation}
where $ \lambda_c $ and $ \lambda_f $ balance the contributions of the coarse and fine level supervision. $\mathbf{w}^x$ and $\mathbf{w}^y$ denote the parameters of the fine-level regression heads along the horizontal $x$ and vertical $y$ axes.

\section{EXPERIMENTS}
\begin{table*}[t]
\caption{Results of Relative Pose Estimation on ScanNet and MegaDepth Dataset. We report the pose estimation AUC under three error thresholds, along with the overall runtime efficiency of the models.}
\centering  
\small
\begin{tabular}{l l c c c c c c c}
\toprule
\multirow{3}{*}{\textbf{Category}} & \multirow{3}{*}{\textbf{Method}} & \multicolumn{3}{c}{\textbf{ScanNet}} & \multicolumn{3}{c}{\textbf{MegaDepth}} & \multirow{3}{*}{\textbf{Time(ms)}}\\
\cmidrule(r){3-5} \cmidrule(r){6-8}
& & AUC@5° & AUC@10° & AUC@20° & AUC@5° & AUC@10° & AUC@20° \\
\midrule
\multirow{3}{*}{Sparse} 
& SP + SG~\textsubscript{\textit{\scriptsize CVPR'20}} & 16.2 & 32.8 & 49.7 & 57.6 & 72.6 & 83.5 & 96.9 \\
& SP + LG~\textsubscript{\textit{\scriptsize ICCV'23}} & 14.8 & 30.8 & 47.5 & 58.8 & 73.6 & 84.1 & 84.2\\
& XFeat~\textsubscript{\textit{\scriptsize CVPR'24}} & 16.7 & 32.6 & 47.8 & 44.2 & 58.2 & 69.2 & 14.2\\

\midrule
\multirow{2}{*}{Dense} 
& DKM~\textsubscript{\textit{\scriptsize CVPR'23}} & 26.6 & 47.1 & 64.1 & 67.3 & 79.7 & 88.1 & 554.2\\
& ROMA~\textsubscript{\textit{\scriptsize CVPR'24}} & 28.9 & 50.4 & 68.3 & 68.5 & 80.6 & 88.8 & 824.9\\
\midrule
\multirow{7}{*}{Semi-Dense} 
& LoFTR~\textsubscript{\textit{\scriptsize CVPR'21}} & 16.9 & 33.6 & 50.6 & 62.1 & 75.5 & 84.9 &117.5\\
& MatchFormer~\textsubscript{\textit{\scriptsize ACCV'22}} & 15.8 & 32.0 & 48.0 & 62.0 & 75.6 & 84.9 & 156.0\\
& ASpanFormer~\textsubscript{\textit{\scriptsize ECCV'22}} & 19.6 & 37.7 & 54.4 & 62.6 & 76.1 & 85.7 & 155.7\\
& RCM~\textsubscript{\textit{\scriptsize ECCV'24}} & 17.3 & 34.6 & 52.1 & 58.3 & 72.8 & 83.5 & 93.0\\
& E-LoFTR~\textsubscript{\textit{\scriptsize CVPR'24}} & 19.2 & 37.0 & 53.6 & 63.7 & 77.0 & 86.4 & 69.6\\
& JamMa~\textsubscript{\textit{\scriptsize CVPR'25}} & 15.1 & 31.6 & 48.5 & 64.1 & 77.4 & 86.5 & \textbf{59.9}\\
& \textbf{SURE (Proposed)} & \textbf{20.3} & \textbf{38.6} & \textbf{55.3} & \textbf{64.7} & \textbf{77.7} & \textbf{86.8} & 62.8\\
\bottomrule
\end{tabular}

\vspace{2mm}

\label{tab:relative_pose}
\end{table*}





\subsection{Implementation Details}
Our model is trained on the MegaDepth dataset~\cite{li2018megadepth}. We adopt the standard training and testing splits provided by~\cite{wang2024efficient} to ensure fair comparisons with existing methods. 
The entire pipeline is trained end-to-end using the AdamW optimizer, with an initial learning rate of $2 \times 10^{-3}$ and a weight decay of $1 \times 10^{-4}$. 
In our implementation, the fine-level loss weight $\zeta$ is fixed at 1, while the focal loss parameters $\gamma$ and $\alpha$ are set to 2 and 0.25, respectively.
The total loss is formulated as a weighted sum of the objectives of the coarse level and the fine level, with respective weights set to $\lambda_c = 1.0$ and $\lambda_f = 0.25$. We train the model for 30 epochs using 4 NVIDIA 3090 GPUs, with a batch size of 16 and an input image resolution of $832 \times 832$.
The bin number N set to 16. The temperature parameter $\tau$ used in softmax normalization is fixed to 0.1 throughout training. The trustworthiness regression network consists of two 1D convolutional layers. During inference, the uncertainty thresholds for filtering, $\tau_a$ and $\tau_e$, are both set to 0.95.

\subsection{Relative Pose Estimation}\label{sec:evaluation}

\subsubsection{Dataset}
Following prior protocols~\cite{sun2021loftr}, we evaluate on 1500 image pairs each from ScanNet~\cite{dai2017scannet} (indoor) and MegaDepth~\cite{li2018megadepth} (outdoor). All ScanNet images are resized to 640×480, while MegaDepth images are uniformly scaled to 832×832 for consistency


\subsubsection{Baselines} 
We compare with three categories of methods: (1) sparse keypoint-based matchers including SuperPoint~\cite{detone2018superpoint}, SuperGlue~\cite{sarlin2020superglue},  LightGlue~\cite{lindenberger2023lightglue}, and XFeat~\cite{potje2024cvpr}; (2) semi-dense matchers such as RCM~\cite{lu2025raising}, LoFTR~\cite{sun2021loftr}, MatchFormer~\cite{wang2022matchformer}, AspanFormer~\cite{chen2022aspanformer}, E-LoFTR~\cite{wang2024efficient},and JAMMA~\cite{lu2025jamma}; (3) dense matcher ROMA~\cite{edstedt2024roma} and DKM~\cite{edstedt2023dkm}.

\subsubsection{Evaluation Protocol}
Following previous work~\cite{lu2025jamma}, we report the area under the curve (AUC) of the pose error at thresholds of $5^\circ$, $10^\circ$, and $20^\circ$. We use LO-RANSAC to estimate the essential matrix for MegaDepth, and RANSAC for ScanNet. All methods use a RANSAC inlier threshold of $0.5$. Furthermore, we measure the matching runtime on the MegaDepth dataset using a single NVIDIA 4090 GPU to assess efficiency.

\subsubsection{Results}
As shown in Tab.~\ref{tab:relative_pose}, our method achieves state-of-the-art performance among sparse and semi-dense matchers on both MegaDepth and ScanNet. It surpasses recent semi-dense approaches such as E-LoFTR, ASpanFormer and JamMa across all AUC thresholds, while maintaining high computational efficiency. Compared to dense matchers like DKM and RoMa, our model achieves a more favorable balance between accuracy and speed, making it more suitable for real-time applications. As illustrated in Fig.~\ref{fig3}, we visualize the correspondences along with their estimated uncertainties, demonstrating that our model produces cleaner and more reliable matches overall. 

As further evidence, Fig.~\ref{fig4} highlights how our uncertainty modeling behaves in challenging scenarios. Out of 2048 predicted correspondences, we visualize the 100 pairs with the highest uncertainty. We observe that epistemic uncertainty tends to localize occluded areas under large viewpoint changes, while aleatoric uncertainty concentrates on weak-texture regions. Since these areas are also prone to incorrect matches, the results confirm that our uncertainty estimation is both meaningful and effective. By filtering or down-weighting such matches, our method not only improves robustness but also prevents error propagation to downstream tasks.

\subsection{Homography Estimation}

\subsubsection{Dataset} We conduct evaluation on the HPatches benchmark~\cite{balntas2017hpatches}, which has 108 planar image sequences.

\subsubsection{Baselines}
We compare our method against recent sparse and semi-dense approaches. Sparse matchers include SuperPoint~\cite{detone2018superpoint}, R2D2~\cite{r2d2}, DISK~\cite{tyszkiewicz2020disk}, SuperGlue~\cite{sarlin2020superglue}, and LightGlue~\cite{lindenberger2023lightglue}. Semi-dense matchers include DRC-Net~\cite{li20dualrc}, Patch2Pix~\cite{zhou2021patch2pix}, LoFTR~\cite{sun2021loftr}, ASpanFormer~\cite{chen2022aspanformer}, and E-LoFTR~\cite{wang2024efficient}.

\subsubsection{Metric} Following the protocol in~\cite{sun2021loftr}, we use the top 1000 matches for homography estimation and apply vanilla RANSAC. The AUC of the projection error is calculated at thresholds $3$, $5$, and $10$ pixels. All images are resized so that the shorter side is 480 pixels.

\subsubsection{Results} 
As shown in Tab.~\ref{tab:hpatches}, SURE achieves the best AUC at both 5px and 10px, demonstrating a strong coarse-level localization.
While slightly behind others at stricter thresholds, it is mainly due to the use of more  complex fine matching modules in  existing semi-dense methods. Nonetheless, SURE strikes a favorable balance between accuracy and efficiency.

\begin{table}[t]
\caption{Homography estimation results on the HPatches dataset. We report the AUC of corner reprojection error.}
\centering
\begin{tabular}{llccc}
\toprule
\multirow{2}{*}{Category} & \multirow{2}{*}{Method} & \multicolumn{3}{c}{AUC} \\
\cmidrule(lr){3-5}
& & @3px & @5px & @10px \\
\midrule
\multirow{5}{*}{Sparse} 
& SP + NN & 41.6 & 55.8 & 71.7 \\
& R2D2 + NN & 50.6 & 63.9 & 76.8 \\
& DISK + NN & 52.3 & 64.9 & 78.9 \\
& SP + SG & 53.9 & 68.3 & 81.7 \\
& SP + LG & 54.2 & 68.3 & 81.5 \\
\midrule
\multirow{6}{*}{Semi-Dense} 
& DRC-Net & 50.6 & 56.2 & 68.3 \\
& Patch2Pix &  59.3 & 70.6 & 81.2 \\
& LoFTR & 65.9 & 75.6 & 84.6 \\
& ASpanFormer & \textbf{67.4} & 76.9 &  85.6 \\
& E-LoFTR & 66.5 & 76.4 & 85.5 \\
& \textbf{SURE(ours)} & 67.0 & \textbf{77.2} & \textbf{86.1} \\
\bottomrule
\end{tabular}
\label{tab:hpatches}
\end{table}

\subsection{Ablation Study}


We evaluate key design choices via ablation studies (Tab.~\ref{tab:ablation-auctime}).
All models are trained for 30 epochs at 832 resolution and evaluated using the protocol in Sec.~\ref{sec:evaluation}. AUC@10° is reported on MegaDepth and ScanNet, with runtime measured on MegaDepth.
\textit{I}: We start with E-LoFTR as our baseline.
\textit{II}: Replacing the fine matching module with a simple 2-layer 1-D CNN regression head that directly predicts $(x, y)$ coordinates using L2 loss results in a noticeable drop in AUC to ($35.9\%$ / $75.0\%$), along with a reduced inference time of $-11.3$ ms.
\textit{III}: Decomposing the regression into two 1-D CNN heads for $x$ and $y$ slightly improves performance to ($36.3\%$ / $75.8\%$) at $59.0$ ms, suggesting marginal stability benefits from axis-wise modeling.
\textit{IV}: Introducing our spatial fusion module improves generalization, boosting performance to ($37.3\%$ / $76.7\%$) with a moderate increase in time to $62.5$ ms. This highlights the value of integrating fine-level structural cues.
\textit{V}: Applying a KL loss following SimCC~\cite{li2022simcc} does not offer further gains, resulting in ($37.2\%$ / $76.5\%$). This suggests that our proposed uncertainty loss better captures spatial confidence.
\textit{VI}: Replacing the L2 loss with our evidential formulation improves the performance to ($38.2\%$ / $77.3\%$), with negligible time cost ($62.7$ ms), confirming the advantage of uncertainty-aware regression.
\textit{VII}: Filtering unreliable matches using predicted uncertainty yields the best result ($38.6\%$ / $77.7\%$), with a minimal time increase to $62.8$ ms. 
Compared to E-LoFTR, our method delivers both higher accuracy and lower latency, highlighting its superior balance between effectiveness and efficiency.

\begin{table}
\caption{Ablation study on MegaDepth and ScanNet datasets. We report AUC@10° and inference time. Each component is examined to assess its individual and combined impact on performance.}
\centering
\small
\setlength{\tabcolsep}{2.5pt}  
\begin{tabular}{lccc}
\toprule
\multirow{2}{*}{Setup~$\downarrow$} & \multicolumn{2}{c}{AUC@10°~$\uparrow$} & \multirow{2}{*}{Time~(ms)~$\downarrow$} \\
\cmidrule(lr){2-3}
& ScanNet & MegaDepth & \\
\midrule
\textit{I} (Baseline): E-LoFTR & 37.0 & 77.0 & 69.6 \\
\textit{II}: Direct Regression & 35.9 & 75.0 & 58.3 \\
\textit{III}: Coord. Regression & 36.3 & 75.8 & 59.0 \\
\textit{IV}: Spatial Fusion & 37.3 & 76.7 & 62.5 \\
\textit{V}: KL Loss & 37.2 & 76.5 & 62.5 \\
\textit{VI}: Evidential Head & 38.2 & 77.3 & 62.7 \\
\textit{VII} (Ours): Filtering  & \textbf{38.6} & \textbf{77.7} & 62.8 \\
\bottomrule
\end{tabular}
\label{tab:ablation-auctime}
\end{table}



\begin{figure}[t]
  \centering
  \begin{subfigure}[t]{0.22\textwidth}
    \centering
    \includegraphics[width=\linewidth]{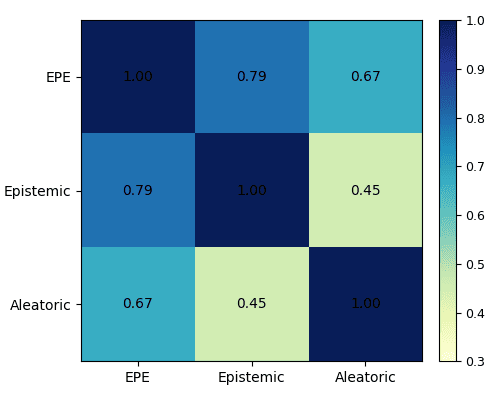}
    \caption{Megadepth }
  \end{subfigure}
  \hfill
  \begin{subfigure}[t]{0.22\textwidth}
    \centering
    \includegraphics[width=\linewidth]{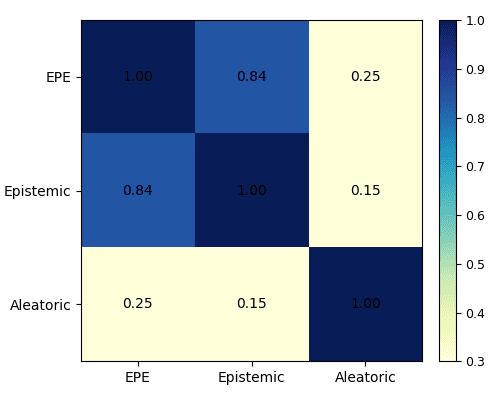}
    \caption{Scannet }
  \end{subfigure}

  \caption{Uncertainty analysis. (a) and (b) are the heat maps of the Spearman rank correlation analysis between the EPE and the uncertainties.}
  \label{fig:uncertainty_heatmaps}
\end{figure}

\subsection{Uncertainty Analysis}
To evaluate uncertainty quality, we compute the Spearman correlation between end-point error (EPE) and two types of uncertainties: aleatoric (data)  uncertainty and epistemic (model)  uncertainty, as shown in Fig.~\ref{fig:uncertainty_heatmaps}. 
In MegaDepth, model uncertainty correlates more strongly with EPE $0.79$ than data uncertainty $0.67$, indicating its superior alignment with actual prediction error. The modest correlation between the two $0.45$ suggests they capture complementary aspects—data uncertainty reflects input noise, while model uncertainty measures confidence due to limited knowledge.
In ScanNet, model uncertainty remains reliable with a higher correlation $0.84$, whereas data uncertainty drops to $0.25$, likely due to domain shift. This highlights the robustness of epistemic uncertainty across datasets, and the sensitivity of aleatoric uncertainty to scene-specific variations.
These results confirm that our uncertainty estimates provide reliable indicators of prediction confidence.
\section{CONCLUSIONS}
We present a novel matching framework that integrates uncertainty-aware learning with an efficient spatial fusion strategy to enhance the accuracy of the correspondence. By jointly modeling confidence and structure, our method effectively captures both semantic context and spatial precision. Extensive experiments on challenging benchmarks demonstrate superior performance, and ablation studies confirm the contribution of each component. These results highlight the importance of incorporating probabilistic reasoning and fine-grained feature cues in semi-dense feature matching.


\bibliographystyle{IEEEtran}
\bibliography{IEEEabrv,reference}

\end{document}